\documentclass[10pt, a4paper]{article}

\usepackage{charter}
\usepackage{a4wide}
\usepackage{natbib}
\usepackage{url}
\setcitestyle{authoryear,open={(},close={)}} 
\usepackage{algorithm,algorithmic}
\usepackage{stmaryrd}
\usepackage{tikz}
\usepackage{bm}
\usepackage{dutchcal}

\usepackage{wrapfig,booktabs}
\usepackage{authblk}

\title{Analogical Proportions and Creativity:\\ A Preliminary Study}

\author[1]{Stergos Afantenos} 
\author[1]{Henri Prade}  
\author[1]{Leonardo Cortez Bernardes}

\affil[1]{IRIT -- CNRS, 118, route de Narbonne, 31062 Toulouse Cedex 9, France \\
    \texttt{stergos.afantenos@irit.fr, prade@irit.fr, leonardo.cortez-bernardes@univ-tlse3.fr}}

\date{}
\begin{document}

\maketitle

\begin{abstract}
Analogical proportions are statements of the form ``$a$ is to $b$ as $c$ is to $d$'', which expresses that the  comparisons of the elements in pair $(a, b)$ and in pair  $(c, d)$ yield similar results. Analogical proportions are creative in the sense that given 3 distinct items, the representation of a 4th item $d$, distinct from the previous items, which forms an analogical proportion with them can be calculated, 
provided certain conditions are met. After providing an introduction to analogical proportions and their properties, the paper reports the results of an experiment  made with a database of animal descriptions and their class, where we try to ``create '' new animals from existing ones, retrieving  rare animals such as platypus. We perform a series of experiments using word embeddings as well as Boolean features in order to propose novel animals based on analogical proportions, showing that word embeddings obtain better results.
\end{abstract}

\section{Introduction}
Creativity has raised interest for a long time in computer sciences and in AI \citep{Boden,Schmidhuber,Colton08} with applications in many areas. 
Analogical reasoning has  always been known to foster creativity, especially in creative thinking and problem solving \citep{HolTha1995,Goel1997design,Veale2006}. Indeed analogical reasoning makes a parallel between two situations, which suggests that what is true or  applicable in the first situation might be true or applicable as well in the second situation which presents some similarity with the first one. 

Analogical proportions \citep{PraRicIJCAI2021} are quaternary relations denoted $a : b :: c : d$ between four items $a, b, c, d$, which read ``$a$ is to $b$ as $c$ is to $d$''. In the following, $a, b, c, d$ are represented by means of vectors ; these vectors may  either be made of feature values, or be word embeddings \citep{MikolovNIPS2013}. Analogical proportions can be viewed as a building block of analogical reasoning. Indeed they draw parallels between the ordered pairs $(a, b)$ and $(c, d)$. For example ``\emph{the calf is to the cow as the foal is to the mare}'' put bovidae on a par with equidae. In such  analogical proportions, the four items can be described by means of {\it the same set of features}. Analogical proportions have been successfully applied to classification  \citep{MicBayDelJAIR2008,BouPraRicIJAR2017b,ijar/BounhasP23}, in preference prediction \citep{aaai/FahandarH18,sum/BounhasPPS19}, or for solving Raven IQ tests \citep{BeltranPR16,RagNeuECAI2014}. 

\citet{MikolovNIPS2013} showed that embeddings language models have the potential to respect analogical proportions in a vector space, although later approaches showed that this was due to the limited corpus of analogies that was used proposing better resources for testing analogies \citep{gladkova-etal-2016-analogy,wijesiriwardene-etal-2023-analogical}. Recently, \citet{hu-etal-2023-context} have showed that Large Language Models have the capacity to solve Raven problems, provided to them in a natural language description, performing at least at the same level as human beings. To our knowledge, 
the creative capacities of analogies to produce something new have not been explored, especially in natural language.

This paper presents an investigation of the creative power of analogical proportions. Their creative power relies on their capabilities to produce a fourth item from three items (provided that some conditions hold) as we shall see. The paper is organized  into two main sections. The first one provides the necessary information on analogical proportions, the second one proposes an experiment based on the Zoo dataset\footnote{https://archive.ics.uci.edu/dataset/111/zoo}, where we ``create'' animal descriptions from existing ones, and see if the results exist or not in the database. We provide two sets of experiments. The first is based on predicate representations, while the second is based on vectorial representations of words. The initial experiments that we present in this paper utilize word embeddings from the GloVe framework \citep{pennington-etal-2014-glove}, obtaining promising results.

\section{Analogical proportions}
This section is structured in five subparts i) recalling the Boolean logical modeling, postulates, and properties of analogical  proportions (AP), ii) providing an example showing their  creative power, iii) handling nominal attribute values, 
 iv) introducing nested APs, and  
v) dealing with word APs. 

\subsubsection*{Truth table, postulates and properties}
A logical modeling of an AP ``$a$ is to $b$ as $c$ is to $d$'' where $a, b, c, d$ are Boolean variables is given by the following formula that expresses that 
$a$ \emph{differs} from $b$ as $c$ \emph{differs} from $d$ and $b$ \emph{differs} from $a$ as $d$ \emph{differs} from $c$'' \citep{MicPraECSQARU2009}: 
\[a:b::c:d = ((a \wedge \neg b)  \equiv (c \wedge \neg d)) \wedge ((\neg a \wedge b)  \equiv (\neg c \wedge d))\]
This expression is true for the 6 patterns given 
in Table  \ref{truthTableAnalogy}  and false for the $2^4 -6 =10$ other possible patterns.  
\begin{table}[ht]
\vspace{-0.5cm}
\centering
$$\begin{array}{ccccc}
a & b & c & d  \\
\hline
\hline
0 & 0 & 0 & 0  \\
\hline
1 & 1 & 1 & 1  \\
\hline
0 & 0 & 1 & 1  \\
\hline
1 & 1 & 0 & 0  \\
\hline
0 & 1 & 0 & 1 \\
\hline
1 & 0 & 1 & 0 \\
\hline
\end{array}
$$
\caption{Boolean patterns making analogical proportion $a:b::c:d$ true}
\label{truthTableAnalogy}
\end{table}

 This is the minimal Boolean model that satisfies the three following basic postulates (inspired from numerical proportions) that an AP should obey \citep{PraRicIJAR2018}:

\begin{itemize}
  \item {\it reflexivity}: $a:b:a:b$ ;
  \item {\it  symmetry}: $a:b::c:d \Rightarrow c:d::a:b$ ;
  
  \item {\it  central permutation}: $a:b::c:d \Rightarrow a:c::b:d$.
\end{itemize}
As a consequence, we have the properties: 
\begin{itemize}
    \item $a:a:b:b$ (identity),
    \item $a:b::c:d \Rightarrow$ $b:a::d:c $ (internal reversal), and
    \item $a:b::c:d \Rightarrow d:b::c:a $ (external permutation).
\end{itemize}
Remarkably enough, Boolean APs are \emph{code independent}, i.e.,    $a:b::c:d \Rightarrow \neg a: \neg b:: \neg c: \neg d$. Thus any property used for describing items can be encoded positively or negatively.

We assume that the items considered are represented by Boolean \emph{vectors} with $n$ components corresponding to $n$ feature values, i.e.,  $\vec{a}\!=\!(a_1, ..., a_n)$, etc.  An analogical proportion ``$\vec{a}$ is to $\vec{b}$ as $\vec{c}$  is to  $\vec{d}$'', denoted ${\vec{a}:  \vec{b}:: \vec{c}: \vec{d}} $, is defined componentwise:
${\vec{a}:  \vec{b}:: \vec{c}: \vec{d}} \mbox{ if and only if } \forall i \in [1,n],  a_i:b_i::c_i:d_i$ 

\subsubsection*{Analogical proportions are creative} 
As an illustration, let us consider the geometric analogy problem in Fig. below \citep{BeltranPR16} \citep{PraRicIJCAI2021}.
\begin{figure}[ht]
\begin{center}
\includegraphics[scale=0.35]{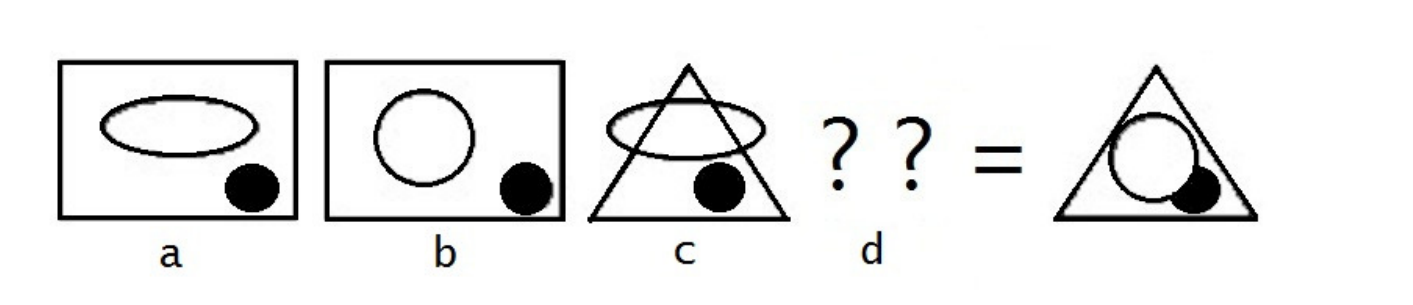}\vspace{-0.9cm}
\label{IQtest}
\end{center}
\end{figure}

It  
 can be encoded with five  Boolean predicates: \textit{hasRectangle (hR), hasBlackDot (hBD),  hasTriangle  (hT),  hasCircle  (hC), hasEllipse (hE)}, in that order in Table \ref{ex},  where $a$, $b$, $c$ are encoded. Each column is an AP equation
 $a_i : b_i :: c_i : x_i$. 
 
 An  equation $a:b::c:x$ \emph{has not always a solution}. Indeed the equations $1:0::0:x$, and $0:1::1:x$ 
do not have a solution so that the analogical proportion holds  
as one of the 6 patterns making an AP true. If  $a:b::c:x$ has a solution, it is  \emph{unique}. 
This is the case in the example,  where $\vec{x}= (0 1 1 1 0)$ in Tab. \ref{ex} 
 to  $\vec{x}=\vec{d}$,  drawn on  the right of Fig. above. 

It should be emphasized that here the description of $\vec{d}$ is computed  directly from those of $\vec{a}, \vec{b}, \vec{c}$, which contrasts with this kind of (easy) IQ tests where the answer is to be found in a set of several candidate solutions. Indeed the first program for solving such problems \citep{Evans68}, in the early years of AI, was selecting the solution among candidate solutions on the basis of a distance accounting for the similarity of transformations for going from $\vec{a}$ to $\vec{b}$ and from $\vec{c}$ to $\vec{x}$ (where $\vec{x}$ is the considered candidate solution). The fact that the description
of $\vec{x}$ is now directly computed from $\vec{a}, \vec{b}, \vec{c}$ shows the {\it creative power} of the analogical proportion. Moreover  {\it $\vec{d}=\vec{x}$ is   distinct from $\vec{a}$,  $\vec{b}$ and  $\vec{c}$}. This is general in $\vec{a} : \vec{b} :: \vec{c} : \vec{d}$ as soon as $\vec{a}$ is distinct from $\vec{b}$ and distinct from $\vec{c}$ (on at least two attributes). 

\begin{table}[ht]
\begin{center}
\begin{tabular}{|c|c|c|c|c|c|}
\hline
$~~~$ & $~hR~$ & $~hBD~$ & $~hT~$ & $~hC~$ & $~hE~$ \\
\hline
$\vec{a}$ & 1 & 1 & 0 & 0 & 1 \\
$\vec{b}$ & 1 & 1 & 0 & 1 & 0 \\
$\vec{c}$ & 0 & 1 & 1 & 0 & 1 \\
\hline
$\vec{x}$ & ? & ? & ? & ? & ?  \\
\hline
\end{tabular}
\caption{Encoding the example. $\vec{x}= (0 1 1 1 0)$ }
\label{ex}
\end{center}
\end{table}

\subsubsection*{Nominal values} 
The description of items may involve nominal attributes, i.e., attributes with a finite domain whose cardinality larger than 2. Then $a:b::c:d$ holds  for nominal variables if and only if
\citep{PirYvon99},  
\citep{BouPraRicIJAR2017b}: 

\[(a, b,  c, d) \in \{{(s, s, s, s)}, {(s, t, s, t)}, { (s, s, t, t)} | \ s, t \in \mathcal{A}\}\]
 
\noindent where $s$, $t$  stand for any value of the attribute domain $\mathcal{A}$. It  generalizes the Boolean case and preserves all the properties reported for this case. 

\subsubsection*{An example of {\it nested} analogical proportion} 
We now illustrate nominal values with an example of AP (from  \citep{Klein1982}) between four sentences ${\bm a}=$ ``girls hate light'', ${\bm b}=$ ``boys love light'', ${\bm c}=$ ``women hate dark'', ${\bm d}=$ ``men love dark'', 
viewed as ordered tuples of nominal values, with respect to the 3 attributes `subject', `verb', `complement'. Each sentence is thus  a vector with 3 components:  
\begin{table}[ht]
\begin{center}
\begin{tabular}{|c|ccc|}
\hline
& subject & verb  & complement\\
\hline
$\vec{a}$ & girls & hate & light  \\
$\vec{b}$ & boys & love & light  \\
$\vec{c}$ & women & hate & dark \\
$\vec{d}$ & men & love & dark   \\
\hline
\end{tabular} 
\caption{Analogical proportion between sentences}
\label{Klein} 
\end{center}
\vspace{-0.6cm}
\end{table}

Obviously, the values in the  `subject' column of  Table \ref{Klein} do  not make an analogical proportion by themselves, since   four distinct values are involved (while $\texttt{hate} : \texttt{love} :: \texttt{hate} : \texttt{love}$, and $\texttt{light} : \texttt{light} :: \texttt{dark} : \texttt{dark}$ are clear analogical proportions in terms of nominal values). However $\texttt{girls} : \texttt{boys} :: \texttt{women} : \texttt{men}$ can be viewed as a compact writing of an AP between descriptions in terms of a collection of Boolean features as in Table \ref{Klein2}, where an AP holds in each column. So $\vec{a} :  \vec{b} :: \vec{c} : \vec{d}$, and thus  ${\bm a}: {\bm b} :: {\bm c} : {\bm d}$ do hold. The AP $\vec{a} :  \vec{b} :: \vec{c} : \vec{d}$ is a {\it nested} one since it involves an attribute whose four  values do not form one of the three basic patterns of a  nominal analogical proportion, while those values are associated with vector descriptions which themselves form an AP.

\begin{table}[!ht]
\centering
\begin{tabular}{|c|ccccc|}
\hline
& \emph{male} & \emph{female}  & \emph{young} & \emph{adult} & \emph{human} \\
\hline
\texttt{girls}  & 0 & 1 & 1& 0 & 1  \\
\texttt{boys}  & 1 & 0 & 1 & 0 & 1 \\
\texttt{women} & 0 & 1 & 0 & 1 & 1  \\
\texttt{men}  & 1& 0  & 0 & 1 & 1 \\
\hline
\end{tabular}
\caption{Girls are to  boys as   women  are to   men}\label{Klein2}
\end{table}
\subsubsection*{Word analogical proportions}
In the example of Table \ref{Klein}, the values of the vector components are words. Words 
can themselves be represented by vector embeddings  \citep{MikolovNIPS2013}. Analogical proportions can be directly defined in terms of such vector representations, as early foreseen in 
\citep{RumelhartAbrahamson1973}. Then the analogical proportion $\vec{a} :  \vec{b} :: \vec{c} : \vec{d}$ holds if and only if $\vec{a} -  \vec{b} = \vec{c} - \vec{d}$, i.e.,  $\forall i, a_i -b_i= c_i -d_i$. Note that this agrees with the truth Table of page~\pageref{truthTableAnalogy} in the Boolean case: when $a_i, b_i, c_i, d_i \in \{0, 1\}, a_i  : b_i :: c_i : d_i \Leftrightarrow a_i -b_i= c_i -d_i $. See  
\citep{aicom/PradeR21} for more details on this view. Thus, the vector embeddings of words `girls', `boys', `women', and `men' should be such that 
$\vec{v_{girls}}- \vec{v_{boys}}= \vec{v_{women}}- \vec{v_{men}}$. 

Moreover this view agrees with the  computation of vector embeddings of sentences as the sum of the vector embeddings of the words in the sentence. Indeed we have $(\vec{a} -  \vec{b}) + (\vec{a'} -  \vec{b'}) = (\vec{c} - \vec{d}) + (\vec{c'} - \vec{d'}) \Leftrightarrow (\vec{a} + \vec{a'}) - (\vec{b} + \vec{b'}) = (\vec{c} + \vec{c'}) - (\vec{d} + \vec{d'})$, which means that if two 4-tuples $(\vec{a}, \vec{b}, \vec{c}, \vec{d})$ and $(\vec{a'}, \vec{b'}, \vec{c'}, \vec{d'})$ of words are an AP, their additive grouping is also an AP.
Indeed, in order to confirm that, we used GloVe vector embeddings representing each sentence as the mean of the embedding vectors of its words. We then calculated
$1 - \cos\frac{\vec{b} - \vec{a}}{\vec{d} - \vec{c}}$
which represents how close the vectors are in forming an analogy, obtaining   a value of $0.8513$ showing that the 4 vectors are close to forming a parallelogram. 
\vspace{-0.2cm}
\section{Experiments} 

{\bf Procedure}. We start with a database containing the descriptions of animals in terms of Boolean features. We also assume that a class is known for each animal. Thus, an animal $A$ is described by a vector $\vec{a}$ together with its class $cl(\vec{a})$. The idea, given a subset $S$ of the database, is to compute the solutions of equations of the form $\vec{a} : \vec{b}:: \vec{c} : \vec{x}$ where 
$ \vec{a},  \vec{b}, \vec{c}\in S $, and to see if $\vec{x}$ is, or not, in the database. Also we may use only the restriction of vectors $\vec{a}, \vec{b}, \vec{c}$
to a subset of features. More precisely, we look for small subsets of features used for the description of animals, as in the example of 
Table \ref{platypus} (where the solution indeed exists in the animal reign).

\begin{table}[ht]
\begin{center}
\begin{tabular}{|c|c|c|}
\hline
$~~~$ & $~suckle~their~young~$ & $~lay~ eggs~$  \\
\hline
${scorpions}$ & 0 & 0  \\
${ mammals}$ & 1 & 0  \\
${ birds}$ & 0 & 1 \\
\hline
${\bm ?}$ & 1& 1 \\
\hline
\end{tabular}
\end{center}
\vspace{-0.3cm}
\caption{\quad ${\bm ?}$ = {\it  monotremes}}
\label{platypus}
\vspace{-0.4cm}
\end{table}

In this example, the analogical equation solving ``creates'' an animal species  that both lays eggs and suckles her young. It turns out that such animals exist: platypus, echidnas. 

Note that when looking for analogical solutions, we should take care of two issues: 1; the equation should be solvable for each feature considered; 2. $\vec{a} : \vec{b}:: \vec{c} : \vec{x}$ and $\vec{a} : \vec{c}:: \vec{b} : \vec{x}$ have the same solution if any. Moreover, from a creativity point of view the  vectors
$\vec{a},  \vec{b}, \vec{c}$
(or the sub-vectors used) refer to animals that should be sufficiently different. This is why it may be useful to enforce  the constraints $cl(\vec{a})\neq  cl(\vec{b}), cl(\vec{a})\neq cl(\vec{c}), cl(\vec{b})\neq  cl(\vec{c})$.

Besides, we are also interested in solving analogical equations when vectors are word embeddings, i.e., computing  $\vec{x}= \vec{c} + \vec{b} - \vec{a}$, and looking for the words whose embedding is close to $\vec{x}$. The question is then to see if the results are compatible with those obtained from the Boolean representations.

\subsubsection*{Implementation} 
We use the {\it Zoo} dataset (see footnote 1).  This dataset contains 101 animals each one described by 16 features and their class. Out of these 17 features, 15 are binary while two are nominal. The first nominal feature describes the number of legs that the animal has while the other one describes its class (mammal, reptile, etc).  

Initial experiments took into account the full gamut of 16 features, excluding the class of the animal. For the full list of 101 animals, we calculated all possible triplets $101 \choose 3$ and kept only the ones for which the three animals belonged to a different class. For each triplet $(\vec{a},\vec{b},\vec{c})$ we predicted the existence, or not, of a fourth element representing a ``potential'' animal using analogical proportions for each feature, as described in the previous sections. 

Imposing analogical proportions for the full 16 features is quite restrictive, since it suffices for one feature to not be in AP in order to discard the whole triplet. Examining all possible subsets of features of cardinality 2 or more is computationally prohibitive since this would result in ${101 \choose 3}\times 2^{15}$ instances. We wanted thus to identify the subset of the most \emph{``important''} features in order to perform our experiments. In order to do so, we identified the features which had values shared by most animals and we selected the top five. These features were \texttt{hair, eggs, milk, venomous} and \texttt{domestic}. We applied the same procedure as before for this subset of features. 


In  further experiments we investigated how embeddings representing words can cope with creativity using analogies. As explained above, each element in a triplet $(\vec{a},\vec{b},\vec{c})$ is represented in a vectorial form with real values. We predict a 4th element $\vec{d}$ such that $\vec{d}= \vec{c} + \vec{b} - \vec{a}$. We represent each animal $A, B, C$ using 
GloVe \citep{pennington-etal-2014-glove} embeddings with 300 dimensions (``common crawl'' with 840 billion tokens).
After calculating $\vec{d}$ we find the words represented by the closest Glove embeddings to this vector using  Euclidean distance. 

\subsubsection*{Results and discussion} 
Evaluating creativity is still an open research subject. There are no widely accepted measures of creativity and most evaluations remain subjective. In this paper we wanted to have a rough estimate of how often the animals that we proposed did indeed exist in the microworld of the Zoo database. We thus chose to measure the Precision of our results: number of predictions for which at least one animal exists in the Zoo database over all our predictions. The results for the Boolean vectors are shown in Table~\ref{tab:results}.  

\begin{table}[htb]
    \centering
    \begin{tabular}{c|c}
         16 features & 5 most important features \\
         \hline
        22.64\% & 59.67\%
    \end{tabular}
    \caption{Precision results using Boolean vectors.}
    \label{tab:results}
\end{table}

As we can see when we use the full list of 16 features 22.64\% of the proposed animals already exist in the Zoo database while for the 5 most important features 59.67\% exist. Having in mind the subjectivity in evaluating creativity, let us examine some specific cases. Take for example the following three animals: \textit{seasnake, frog, aardvark}. Their vectors for the subset of important features are the following $\vec{a} = (0, 0, 0, 1, 0), \vec{b} = (0, 1, 0, 1, 0), \vec{c} = (1, 0, 1, 0, 0)$ respectively. When we apply the APs on these vectors we obtain vector $\vec{d} = (1, 1, 1, 0, 0)$, predicting thus a hairy animal that lays eggs and milks their children. Although it would seem strange at first, such an animal indeed exists: the \textit{platypus}. Let us consider now the triplet of animals (\textit{platypus, antelope, stingray}) with corresponding vectors: $\vec{a} = (1, 1, 1, 0, 0), \vec{b} = (1, 0, 1, 0, 0), \vec{c} = (0, 1, 0, 1, 0)$. Applying the same procedure we obtain $\vec{d} = (0, 0, 0, 1, 0)$ which corresponds to a non-hairy venomous being that does not lay eggs (when giving birth) and does not milk their children. Such animals do exist: \textit{scorpions} and \textit{seasnakes}. Our approach does indeed identify both of them. 

Regarding GloVe, the computational cost was prohibitive for calculating $101 \choose 3$ instances. Since our goal in this paper is to explore the potential of analogies for creativity, we wanted to compare results between a predicate logic approach and one based on word embeddings. We thus decided to evaluate GloVe on the results that were obtained with the predicate logic based approach and see whether GloVe could match them. After calculating $\vec{d}= \vec{c} + \vec{b} - \vec{a}$ we look for the 10 closest GloVe vectors to $\vec{d}$ according to the Euclidean distance. We discard instances that do not correspond to animals according to WordNet Synsets \citep{mccrae-etal-2019-english}, as implemented by NLTK. We then calculate the precision at $k$, that is for the first $k$ propositions of GloVe we examine whether at least one is present as an animal in the Zoo database, after stemming with NLTK. The results for $k \in [1,8]$ are shown in Table~\ref{tab:results_glove}. As we can see, already for $k=2$ GloVe achieves a precision of 64.60 exceeding the results obtained in the predicate-logic based method, while for $P@8=83.44$.

\begin{table}[htb]
    \centering
    \begin{tabular}{c|c|c|c|c|c|c|c}
        \toprule
        P@1 & P@2 & P@3 & P@4 & P@5 & P@6 & P@7 & P@8\\
        \midrule
        46.33 & 64.60 & 73.94 & 78.75 & 81.33 & 82.60 & 83.21 & 83.44\\
        \bottomrule
    \end{tabular}
    \caption{Precision@k for GloVe vectors  in percents.}
    \label{tab:results_glove}
\end{table}

\section{Concluding remarks} 
This paper has reported a preliminary study on the creative power of APs, showing encouraging results.  In her pioneering work \citet{Boden} distinguishes between three forms of creativity: \emph{combinational, exploratory}, and \emph{transformational}. Combinational creativity is the result of combination of familiar ideas. The other two types of creativity, according to Boden, presuppose the existence of a conceptual space via which new ideas are realized. Thus, in the second type of creativity, the creative agent explores the various corners of the conceptual space seeking instances that can be deemed creative. The final type of creativity presupposes that the agent ``expands'', so to speak, the boundaries of the conceptual space, allowing for new dimensions and thus a deeper understanding of the world and thus the proposal of new creative ideas. 
The creative process that we present in this paper, we argue, is of the second type. In other words, we explore the conceptual space in the microworld of the Zoo database and propose novel instances that have not existed before. 
%
The results show that our proposed method is able to identify the existence of animals that one would not think existed, such as animals that lay eggs and milk their children (platypus) or ones that ovoviviparously give birth to their children but do not milk them (scorpions). 


As far as GloVe results are concerned, we have seen our approach is able to propose ``new'' animals with up to 83.44\% $P@8$ of the proposed animals in the original database, while even results of $P@2$ exceed the Boolean approach. Those results are to be taken only as an indication of the potential of this approach. As \citep{gladkova-etal-2016-analogy} has shown, GloVe works really well when it comes to simple analogy datasets, such the one proposed by \citep{MikolovNIPS2013}, but only 30\% of the analogies are captured in their new BATS corpus. Recent evidence \citep{Webb2023} show that more advanced models based on Transformer technologies are capable of identifying analogies in a manner that is comparable to humans. 

Much remains to be done for understanding computational creativity using APs for the generation of novel creative instances. First of all, we need better measures of the evaluation of creativity which do not take into account only the existence or not of the proposed instances,\footnote{In the experiment with the Zoo dataset, we benefited from the knowledge of existing animals, but such information is rarely available for judging creativity.
} but also somehow measures how ``interesting'' proposed instances are. In the future we plan to work on such measures as well as explore other embedding methods including contextual vectors from BERT \citep{devlin-etal-2019-bert} and other Transformer-based architectures such as GPT-3 and GPT-4. For our experiments with Transformer based architectures we plan to provide full descriptions for animals and their characteristics (or indeed for other concepts) and produce new animals with full descriptions, respecting analogical proportions. Evaluation can be performed by examining whether the proposed concepts do indeed represent animals (using for example Wordnet synsets). Initial experiments concerning the evaluation of the ``creative'' part of the proposed animals could be performed based on the frequency of the combination of their characteristics. More precisely, the rarer the combination of their characteristics is, while still a valid animal, the more creative, one can argue, the proposed animal is. 

\subsubsection*{Acknowledgements} 
This research was supported by the ANR project ``Analogies: from theory to tools and applications'' (AT2TA), ANR-22-CE23-002.


\bibliography{main}

\end{document}